%% file: ms.tex
\documentclass[11pt,a4paper]{article}
\usepackage[hyperref]{acl_format/acl2021}
\usepackage{times}
\usepackage{latexsym}
\usepackage{amssymb}
\usepackage{listings}
\usepackage[flushleft]{threeparttable}
\usepackage{filecontents}
\usepackage{xcolor}

\usepackage{url}

\usepackage[subtle]{savetrees}

\usepackage{booktabs}
\usepackage{graphicx}
\usepackage{caption}
\usepackage{pythonhighlight}
\usepackage{wasysym}

\usepackage{microtype}

\aclfinalcopy 

\newcommand\knodle{\textit{Knodle}}  

\title{Knodle: Modular Weakly Supervised Learning with PyTorch}

\author{Anastasiia Sedova \\
  University of Vienna \\
  Vienna, Austria \\
  \texttt{\normalsize{anastasiia.sedova@univie.ac.at}} \\\And
  Andreas Stephan \\
  University of Vienna \\
  Vienna, Austria \\
  \texttt{\normalsize{andreas.stephan@univie.ac.at}} \\\AND
  Marina Speranskaya \\
  Ludwig Maximilian University of Munich \\
  Munich, Germany \\
  \texttt{\normalsize{speranskaya@cis.lmu.de}} \\\And
  Benjamin Roth \\
  University of Vienna \\
  Vienna, Austria\\
  \texttt{\normalsize{benjamin.roth@univie.ac.at}} \\
  }

\date{}

\begin{document}
\maketitle

\input{chapters/0_abstract}

\input{chapters/1_intro}

\input{chapters/2_related_work.tex}

\input{chapters/3_model.tex}

\input{chapters/4_trainers.tex}

\input{chapters/5_data.tex}

\input{chapters/6_experimental_setup.tex}

\input{chapters/7_conclusion.tex}

\input{chapters/8_acknowledgements.tex}

\bibliographystyle{acl_format/acl_natbib}
\bibliography{bibl}

\end{document}

%% file: chapters/0_abstract.tex
\begin{abstract}
Strategies for improving the training and prediction quality of weakly supervised machine learning models vary in how much they are tailored to a specific task or integrated with a specific model architecture.
In this work, we introduce \mbox{\knodle{}}, a software framework that treats weak data annotations, deep learning models, and methods for improving weakly supervised training as separate, modular components.
This modularization gives the training process access to fine-grained information such as data set characteristics, matches of heuristic rules, or elements of the deep learning model ultimately used for prediction.
Hence, our framework can encompass a wide range of training methods for improving weak supervision, ranging from methods that only look at correlations of rules and output classes
(independently of the machine learning model trained with the resulting labels),
to those that harness the interplay of neural networks and weakly labeled data.
We illustrate the benchmarking potential of the framework with a performance comparison of several reference implementations on a selection of datasets that are already available in \mbox{\knodle{}}.
\end{abstract}

%% file: chapters/1_intro.tex
\section{Introduction}

Most of today's machine learning success stories are built on top of huge labeled data sets.
Creating and maintaining such data sources manually is a time-consuming, complicated and thus an expensive and error-prone process.
Various research directions address the hunger for bigger and better datasets.

One of the most popular approaches that has recently gained traction is \textit{weak supervision}.
The learning algorithm is confronted with labels which are easy to obtain but are not guaranteed to be correct, and as such often demand denoising.
Such weak labels are created, for example, with the use of regular expressions, keyword lists or external databases.
Typically, methods for improving weakly supervised learning (and their respective implementations) are tailored towards domain-specific tasks or integrated with a specific model architecture.
Examples include the attention-over-instances architecture introduced for relation extraction~\cite{lin-etal-2016-neural}, an EM-based algorithm used for event extraction~\cite{o_connor_police} or models of systematic label flips for named entity recognition~\cite{hedderich}.
Such diversity and specificity of approaches makes it difficult to compare or transfer them across tasks without extensive adjustments dictated by the implementation, the task or the data set.

We introduce \knodle{}: a framework for \textit{Kno}wledge-supervised \textit{D}eep \textit{Le}arning, i.e weak supervision with neural networks.
The framework provides a simple tensor-driven abstraction based on PyTorch allowing researchers to efficiently develop methods for improving weakly supervised machine learning models and try them interchangeably to find the one that works the best for a given task.
Within this work, we refer to a denoising method as any method that helps to improve weakly supervised learning regardless the type of noise or bias and the exact level of denoising (weak labels, weak rules etc).

The following points summarize \knodle{}'s main design goals:
\begin{itemize}
\setlength\itemsep{0.5em}
    \item{\textbf{Data abstraction.} A tensor-driven data abstraction subsumes a large number of input variants and is applicable to a diverse range of tasks.}
    \item{\textbf{Method independence.} A decoupled implementation of weak supervision denoising methods and prediction models enables comparability and accounts for domain-specific inductive biases.}
    \item{\textbf{Accessibility.} A high-level interface makes it easy to test existing methods, incorporate new ones and benchmark them against each other.}
\end{itemize}

Several denoising algorithms are already included in \knodle{}.
We also propose a new denoising algorithm, \mbox{\textit{WSCrossWeigh}}, which extends \mbox{\textit{CrossWeigh}}~\citep{wang-etal-2019-crossweigh}, a method for detecting mistakes in crowd-sourced annotation, to the weak supervision setting.
The experiments demonstrate that it outperforms other existing methods on the majority of dataset s.

All implemented methods are tested on several datasets, also included in the \knodle{} ecosystem, and we discuss their performance.
Each dataset exhibits different characteristics,
such as the amount or the precision-recall balance of the used rules.
Moreover, depending on the weakly labeled data set, methods for improving weak labels need to remove spurious matches in some cases, or generalize from them in others.

It is clear that such a diverse problem space should be paired with a rich pool of methods so that the most appropriate denoising method can be found for any task or dataset.
\knodle{} allows to easily explore the spaces of weakly supervised learning settings and label improvement algorithms, and hopefully will facilitate a better understanding of the phenomena that are inherent to weakly supervised learning.

The framework is published as an open-source Python package~\texttt{knodle} and available at~\url{https://github.com/knodle/knodle}.

%% file: chapters/2_related_work.tex
\section{Related work}
\label{sec:related_work}

Many strategies have been introduced to reduce the need for large amounts of manually labeled data.
Among these are \textit{active learning}~\cite{Sun_2012}, where automatically selected instances are manually annotated by experts, and \textit{semi-supervised learning}~\cite{Agichtein_2000,kozareva-etal-2008-semantic}, where a small annotated dataset is combined with a large unlabeled one.
Fine-tuning pretrained language models such as BERT~\cite{devlin2018bert} shows good results if moderate to small amounts of annotations are available.

\subsection{Weak supervision}
\label{subsec:ws}
In weak supervision, tedious expert work is replaced with easy to obtain, but potentially error-prone labels, that are usually derived from a set of heuristic rules.
One of the most popular strategies of weakly supervised learning is \textit{distant supervision}, which uses knowledge from existing data sources to annotate unlabeled data.
The technique is used extensively for relation extraction~\cite{craven,mintz-etal-2009-distant,surdeanu2012multi,riedel-etal-2013-relation,lin-etal-2016-neural}, where various knowledge databases,
such as WordNet~\cite{Snow_2004}, Wikipedia~\cite{Wu_2007} and Freebase~\cite{mintz-etal-2009-distant}, are used as annotation sources.

When using heuristic rules, it is not uncommon that one sample turns out to be annotated by multiple rules.
The most straightforward approach to resolve such cases is majority voting, which is used in early weak supervision algorithms~\cite{protein-protein} as well as in more recent experiments~\cite{Krasakis2019SemisupervisedEL, boland}.
However, majority voting does not deal with the different types of noise introduced by weak supervision, and more noise-specific algorithms are necessary.
For example, the noise produced by \textit{incomplete labels}, which stems from the incompleteness of weak supervision sources and often leads to an increased amount of false negatives,
is commonly reduced by data manipulations, e.g.\ enhancing the knowledge base ~\cite{xu-etal-2013-filling}, a thorough construction of negative examples to balance the positive ones~\cite{riedel-etal-2013-relation}, or explicitly modelling missing knowledge base information with latent variables~\cite{Ritter_2013}.
The problem of \textit{noisy features}, i.e.\ an increased amount of false positive labels stemming from overgeneralization, is often approached by using a relaxed distant supervision assumption~\cite{Riedel_2010,hoffmann-etal-2011-knowledge}, by active learning with additional manual expertise~\cite{Sterckx2014UsingAL}, with help of topic models~\cite{yao-etal-2011-structured, Roth_1013}, as well as by using a combination of multiple methods~\cite{Roth_2014}.

Apart from that, methods treat the identified potentially noisy samples differently.
They are either kept for further training with reduced weights~\cite{jat2018improving,He_2020}, corrected~\cite{shang2019noisy} or eliminated~\cite{qin-etal-2018-robust}.
Thus, denoising methods vary significantly depending on the data and task, what makes the creation of a platform for comparison crucial.

\subsection{Structure Learning}

Structure learning assumes multiple weak labels per instance where each label is created by a so called \textit{labeling function}.
The goal is to learn a dependency structure within these labeling functions which motivates the term structure learning.
Most labeling functions are generated by human intuitions, motivating correlation and dependence between labeling functions.
The first algorithm was implemented in the software package \textit{Snorkel}~\cite{snorkel}, which also implemented the data programming paradigm, allowing to programmatically create labeling functions.
Subsequently improvements were made~\cite{DBLP:journals/corr/BachHRR17, varma2019learning} and variations, such as semi-supdervised learning~\cite{DBLP:journals/corr/abs-1911-09860, maheshwari2020data} were introduced.

\subsection{Noise-aware learning}
\label{subsec:noise-aware learning}

A common idea to mitigate single noisy labels is to build an architecture which accounts for noisy data.
There are different approaches that model noise-robustness by adapting the loss function~\cite{patrini2017making}.
Examples include a generalization of cross-entropy and the mean absolute error ~\cite{zhang2018generalized} or the addition of a special noise layer to a neural network~\cite{sukhbaatar2015training}.
Many approaches are based on noise assumptions, such as on the assumption of symmetric label noise~\cite{vanrooyen2015learning}.
Another approach aims at finding and removing wrongly labeled samples from the training procedure.
An example in this domain is given by the confidence learning framework \textit{CleanLab}, which is based on the intuition that low-confidence predictions in cross-validation are more likely to be labeled wrongly~\cite{northcutt2021confident}.
Note that most of these methods were built with the assumption that there is one label corresponding to each instance, while \knodle{} makes use of several weak signals per instance.

\subsection{Crowdsourcing annotations}
\label{subsec:crowdsourcing}
Another solution to reduce the cost of manual data supervision by experts is \textbf{crowdsourcing}.
In order to increase the supervision accuracy for a task, most crowdsourcing experiments rely on annotations by multiple people, and the final label is defined by majority voting~\cite{Kosinski_2012} or measuring the inter-annotator agreement~\cite{tratz2010taxonomy}.
More sophisticated denoising strategies include anomaly detection~\cite{eskin-2000-detecting}, annotator's reliability modelling~\cite{Dawid_1979}, Bayesian approaches~\cite{Vikas_2012} and generative models~\cite{hovy-etal-2013-learning}.
Some mistakes can be identified by such methods.
For example, mistakes consistently made by careful but biased people~\cite{Ipeirotis_2010}, or errors introduced by \textit{spammers}~\cite{Vikas_2012}.

As both, automatically and human labeled data, are subject to noise and structural errors, many algorithms can be used for both domains.
For example, the MACE algorithm~\cite{hovy-etal-2013-learning}, initially proposed for improving noisy annotations from human annotators, was adapted to the setting of denoising automatically labeled data for named entity recognition~\cite{rehbein-ruppenhofer-2017-detecting}.
With the same motivation, we introduce \textit{WSCrossWeigh} (see Section~\ref{sec:trainers} for more details).
We demonstrate the usefulness of the \textit{Knodle} framework to transfer algorithms for improving crowd-sourced annotations to weak supervision problems.

\subsection{Frameworks}
\label{subsec:frameworks}
\knodle{} is based upon the ideas of several software frameworks.
On a low level, \knodle{} is built on top of PyTorch~\cite{paszke2017automatic}.
As for design decisions, we followed several other high-level libraries that
aim to ease the training and prediction experience.
Namely, we drew inspiration from PyTorch lightning~\cite{falcon2019pytorch}, which in essence tries to remove the burdens of writing your own train loop, and Huggingface's Transformers library~\cite{wolf2020huggingfaces}, which gives easy access to various transformer-based architectures in a fixed manner, so that they can be effortlessly interchanged in code.

%% file: chapters/3_model.tex
\section{Weakly supervised learning with \knodle{}}

\begin{figure*}
\centering
\includegraphics[scale=0.22]{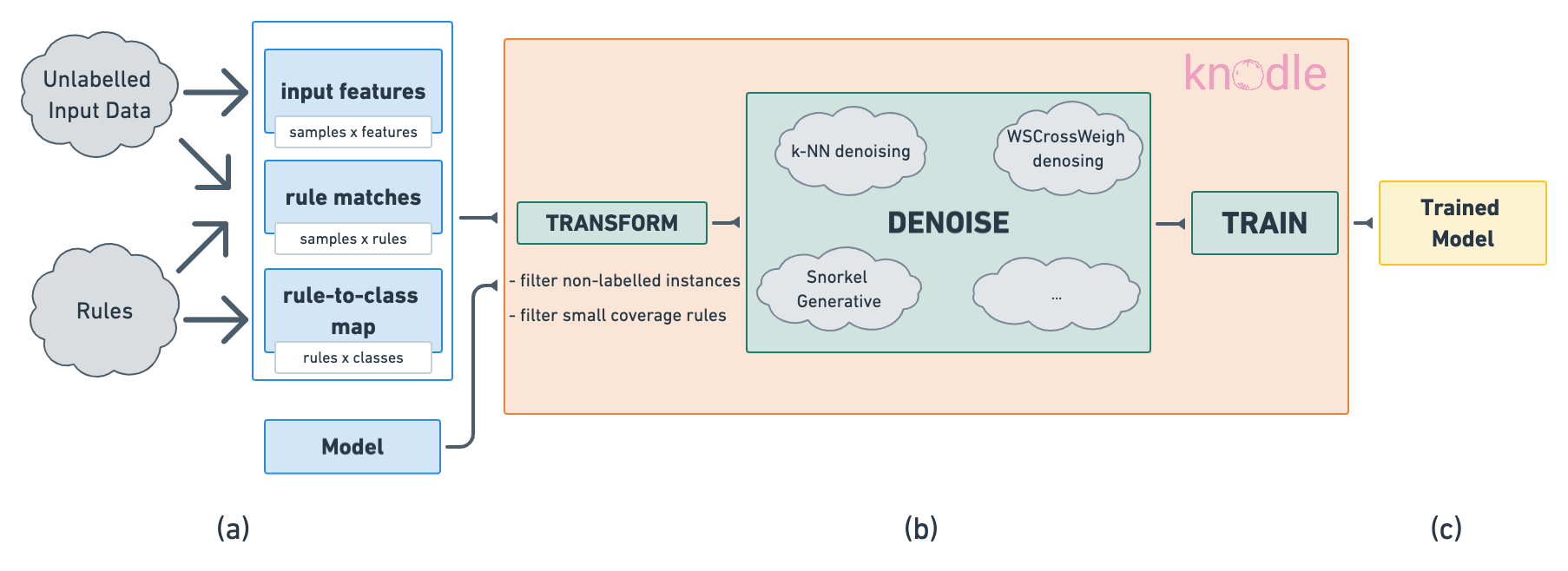} 
\caption{
 	The figure gives an overview of our system. $(a)$ represents the preprocessed input, given as tensors.$(b)$ resembles the internals of \knodle{}.
	The \texttt{Trainer} classes introduced in Section~\ref{subsec:model_impl} handle transformation, denoising and model training. Note that these three steps could be performed subsequently or subsumed in a single training step.  Then, $(c)$ shows the output, a trained PyTorch model.
}
\end{figure*}

The \knodle{} architecture provides a layer of abstraction that allows integrated label improvement and model training with weakly supervised learning signals in PyTorch.
On the one hand, since \knodle{} has access to the information which rules matched for each sample, it is not restricted to methods that denoise 
only weak labels, such as Cleanlab~\cite{northcutt2021confident}.
On the other hand, the \knodle{} abstraction also provides access to input and learned  representations, and thus does not restrict denoising methods to rely on rule match correlations alone (as Snorkel~\cite{snorkel}).
Moreover, access to the deep learning model enables the integration of denoising methods that use or manipulate the prediction model itself.

To the best of our knowledge, \knodle{} is the first framework to provide a modular architecture for interchangable application of a wide spectrum
of denoising algorithms.
For that reason we believe that it can become a testbed where different algorithms for improving the weakly supervised data are implemented and compared with each other to find the most fruitful task-to-denoising-method combination or to use it as a foundation for further studies.

The framework follows two main design principles, outlined below:

\medskip
\textbf{1. Tensor-based representations of input data and weak label matches}
\medskip

Similar to Pytorch models, where the data (input, labels) is already expected to be in tensor format, and the specific pre-processing that led to the tensor representation of the data is outside the scope of the deep learning model implementation, we choose to exclude the process of weak label generation from \knodle{}.
Rather, we encode the information about weak labels in two tensors.
One tensor contains information about which rules matched for each data instance, while another tensor describes the relationship between rules and output classes.

Formally, assume we have $n$ samples, $r$ rules and $k$ classes.
Rule matches are gathered in a binary matrix $Z \in \lbrace 0, 1 \rbrace^{n \times r}$, where $Z_{ij}=1$ if rule $j$ matches sample $i$.
The initial mapping from rules to the corresponding classes is given by another binary matrix $T \in \lbrace 0, 1 \rbrace^{r \times k}$, $T_{jk}=1$ if rule $j$ is indicative of class $k$.

This separation between one tensor that contains rule matches and another tensor that translates them to labels allows \knodle{} to access this fine-grained information during training for certain denoising algorithms.
This is in contrast to other approaches that treat weak supervision as learning from a noisy heuristic label matrix $Y_{heur} = ZT$ without direct access to the individual rules.

\medskip
\textbf{2. Separation of the prediction model from the weak supervision aspects.}
\medskip

\knodle{} requires a standard PyTorch model for a given prediction task.
It is defined independent of the weak supervision aspects, such as rule types or denoising method.
Therefore the same PyTorch model definition can be used for direct or weakly supervised training, and the two settings can easily be compared.
However, even though the prediction model is defined separately, the denoising methods may have access to it during training.
For example, cross-validation schemes such as WSCrossWeigh (see Section~\ref{sec:trainers}) can use the PyTorch model definition for data reweighting or label correction.
This is in contrast to approaches that modularize denoising and training by first adjusting label confidences by using correlations between rules only and then training a model with the adjusted labels~\cite{takamatsu-etal-2012-reducing, snorkel}.
Furthermore, \knodle{}'s design is much more flexible compared to approaches where denoising is so tightly integrated into the underlying prediction model architecture that it could not be changed~\cite{sukhbaatar2015training}.

\subsection{Handling of negative instances}
\label{subsec:neg_samples}
Different tasks need a different logic to handle data samples where no rule matched.
These samples are traditionally called \textit{negative instances}.
Whether unlabeled instances should be used for training (as an additional OTHER class) depends on the task at hand and should be configurable.
For example, in knowledge base population~\cite{Surdeanu} there is only a small number of relevant target relations, and it is important to confidently identify sentences that do not contain any of the target relations (requiring negative instances as examples for the OTHER class).
However, in spam classification with only two classes (spam and not spam) there are rules covering both possible outcomes, and there is no need for unlabeled instances and filtering them out is reasonable.
Current weak supervision frameworks provide only one of the two options: negative samples are either filtered out~\cite{snorkel} or included to the training dataset~\cite{Shu_2020}.

\knodle{} includes configurable functionality for handling such cases (allowing comparability of denoising methods across tasks with and without an OTHER class).
From a technical point of view, there is a \texttt{filter\_non\_labeled} flag in a configuration file, which could be set to \texttt{False} if the negative instances should be filtered out.
To make up for missing explicit annotations for negative samples, an additional \texttt{other\_class} parameter is defined.
Automatically all samples without a matching rule are set to belong to "other" class.
Hence, the exact \texttt{other\_class\_id} could be either provided by the user
or determined automatically by \knodle{}.
These types of configurations are well encapsulated, allowing the specific model to deal with either input.
The amount of negative instances that should included in the training set can be defined specifically for each denoising algorithm.

\subsection{Implementation Details}
\label{subsec:model_impl}

Similar to the most popular deep learning frameworks, such as TensorFlow~\cite{tensorflow2015-whitepaper} and PyTorch~\cite{paszke2017automatic}, we realise learning as a mapping from input tensor(s) to output tensor(s) guided by a loss function that measures the quality of the learned mapping.
However, while the most common solution is to represent the training data by a \emph{design matrix} $X \in \mathbb{R}^{n \times d}$ ($n$ instances represented by $d$ feature dimensions) and a \emph{label matrix} $Y \in \mathbb{R}^{n \times k}$ ($k$ classes), input of \knodle{} are matrices $X$, $Z$ and $T$ described above.
The heuristic labels themselves are calculated later during the weakly supervised learning using the information contained there.
To ensure a seamless use, the weakly supervised algorithms need to be tightly integrated with automatic differentiation and optimization supported by PyTorch.

The denoising and training procedures are realised within \texttt{Trainer} classes.
During initialization, they receive data, a possibly pre-initialized or pretrained model, and a method-specific configuration, inheriting from \texttt{Config} containing information such as model training parameters, criterion, validation method, class weights, various options to handle cases where no rule matches discussed in~\ref{subsec:neg_samples} and others.
The level of integration between denoising and training is different for each \texttt{Trainer}.
Sometimes these procedures can be completely disentangled.
For instance, the \mbox{\texttt{SnorkelTrainer}} firstly denoises the input rules with Snorkel and, secondly, trains the classification model on the purified labels.
Other methods highly integrate denoising and training with each other.
An example is given by the \mbox{\texttt{WSCrossWeighTrainer}}, where several models are trained in oder to calculate sample weights as part of the denoising procedure before the final classifier is trained.

While in standard deep learning frameworks training can be executed by calling \texttt{model.train(X,Y)}, in \knodle{} the same functionality would be invoked with the following command
(illustrates the Trainer with \textit{k}-NN search, which we describe in Section~\ref{sec:trainers}):

\vspace{\baselineskip}
\centerline{\texttt{kNNAggregationTrainer(model, X, Z,}}
\centerline{\texttt{T, config).train()}}
\vspace{\baselineskip}

The following code snippet shows an end-to-end process, starting from data loading, training and evaluation:

\vspace{\baselineskip}
\small{
\begin{lstlisting}[language=python,
				 frame=leftline,keywordstyle=\color{blue},
				 commentstyle=\ttfamily\itshape\color{gray}, showstringspaces=false,
				 xleftmargin=16pt,
				 xrightmargin=3pt,
				 gobble=8,
				 frame=single, basicstyle=\scriptsize\ttfamily,
				 numbers=left]
import torch
from knodle.trainer.knn_aggregation import \
    kNNAggregationTrainer, kNNConfig

# load data in Knodle format
X_train, Z, T, X_test, Y_test = load_data()

# define custom config (or use default)
config = kNNConfig(epochs=2, k=3)

# initialize trainer
trainer = kNNAggregationTrainer(
    model, X_train, Z, T, config
)

# train
trainer.train()

# evaluate
eval_dict = trainer.test(X_test, Y_test)
\end{lstlisting}}
\vspace{\baselineskip}
\normalsize

More detailed information about \mbox{\texttt{kNNAggregationTrainer}} as well as about other \texttt{Trainers} included to \knodle{} is provided in the next section.

%% file: chapters/4_trainers.tex
\section{Trainers}
\label{sec:trainers}

\knodle{} currently provides several out-of-the-box baselines and trainers, which we outline in the following section.
All \texttt{Trainer} classes are compatible with any PyTorch model.
As examples for PyTorch classifiers, \knodle{} provides code using logistic regression and HuggingFace's \texttt{transformers}~\cite{wolf2020huggingfaces}.

\textbf{Majority Voting Baseline.}
As a simple baseline, the rules are directly applied to the test data without any additional model training.
If several rules match, the prediction is done based on the majority;
ties are broken randomly.
As was already mentioned in Section~\ref{sec:related_work}, it is one of the most basic approaches to denoise the data labeled by two or more rules or human annotators.

\textbf{Trainer without Denoising.}
The simplest trained model is the \texttt{NoDenoisingTrainer}.
The majority vote is computed on the training data and used to train the given model.
This is the most direct use of the rule matches for training a classifier.
To cover cases where several rules match, this trainer can be configured to either use a one-hot encoding of the winning label from the majority vote or a distribution over labels (relative to the number of matching rules).

\textbf{Trainer with kNN Denoising.}
This \mbox{\texttt{kNNAggregationTrainer}} includes the label denoising method with a simple geometric interpretation.
The intuition behind it is that similar samples should be activated by the same rules which is allowed by a smoothness assumption on the target space.
The trainer looks at the $k$ most similar samples sorted by, for example, TF-IDF features combined with $L_2$ distance, and activates the rules matching the neighbors to create a denoised $\hat{Z}$.
Importantly, \knodle{} allows separate features for the model training and the neighborhood activation.
This method also provides a way to activate rules for initially unmatched samples.

\textbf{Trainer with Snorkel Denoising.}
\knodle{} provides a wrapper of the Snorkel system~\cite{snorkel} \mbox{\texttt{SnorkelTrainer}}
which incorporates both generative and discriminative Snorkel steps.
The generative step constitutes a denoising method in \knodle's terminology, while the discriminative step corresponds to a prediction model.
The structure within labels and rules, in our notation $P(Y, Z, T)$, is learned in an unsupervised fashion by the generative model.
Afterwards, the final discriminative model, i.e.\ the prediction model, is trained with weak labels provided by the generative model, following the general \knodle{} design.
Both steps are conveniently provided in a single method call.

\textbf{Trainer with Weak Supervision CrossWeigh Denoising.}
Finally, we implemented our own algorithm for noise correction in weakly supervised data.
It is based on the CrossWeight method~\cite{wang-etal-2019-crossweigh} and included to \knodle{} as \mbox{\texttt{WSCrossWeighTrainer}}.
While the original CrossWeigh method was proposed for mistakes identification in crowdworkers annotations, we extend it for denoising the weakly supervised data as well.
In WSCrossWeigh we adopted the same logic for estimating the reliability of weakly annotated data, but made some necessarily corrections specific to weakly supervised learning.

The main intuition behind WSCrossWeigh is the following: if a labeling rule corresponds to a wrong class and, therefore, annotates many samples in the training set with a wrong label, a machine learning model is likely to learn the incorrect pattern and to make similar mistakes when labeling the test samples.
However, if we take a sufficiently big portion of data with samples \textit{not} labeled by this rule, train the model on it, and then classify the samples matched by the rule, the predictions will contradict the initial \textit{wrong} labels, and help us to trace the misclassified samples and reduce their importance in final classifier training.

As in the original CrossWeigh, the basic idea is similar to the \textit{k}-fold cross-validation, where input data is split into \textit{k} folds, each of which becomes, in turn, a test set, while the model is trained on the other folds.
In WSCrossWeigh, however, the splitting is performed not randomly, but based on which rules match for the samples.
Firstly, the rules are randomly split into \textbf{\textit{K}} folds $\{r_1, \dots, r_k\}$ and, iteratively, each $fold_l$ is chosen to form a test set that is built from all samples matched this fold's rules.
Other samples constitute a training set that is used for training the classification model.
During the testing of the trained model on the hold-out fold samples, the predicted label $\hat{y_i}$ for each test sample $x_i$ is compared to the label $y_i$ originally assigned to $x_i$ by weak supervision.
If $\hat{y_i} \neq{y_i}$, this is taken as an indication that the sample $x_i$ is likely to be potentially mislabeled, and its weights $w_{x_i}$ is reduced by a value of an empirically estimated parameter $\epsilon$.
This procedure is repeated several times with different splits to detect misclassified samples more accurately.

The final classifier is trained on the whole reweighed training dataset.
As a result, the more times the original $y_i$ label of data sample $x_i$ was suspected to be wrong, the smaller is its weight $w_{x_i}$, and, therefore, the smaller part it will play in the classifier training.

Along with other denoising algorithms, \mbox{WSCrossWeigh} was tested on the datasets described in Section~\ref{sec:datasets} and showed quite promising results:
it outperforms all other algorithms on three out of four datasets (for more details please see Section~\ref{sec:experiments}).

%% file: chapters/5_data.tex
\section{Datasets}
\label{sec:datasets}

\begin{table*}\centering
    \captionsetup{font=10pt}
   \begin{tabular}{lllrrr}
       \toprule
       \textbf{dataset} & \textbf{classes} & \textbf{train / test samples} & \textbf{rules} & \textbf{avg. rule hits}  & \textbf{class ratio}\\
       \midrule
  Spam &       2 &           1586 / 250 &     10 &            1.63 &       0.47 \\
Spouse &       2 &         22254 / 2701 &      9 &            0.34 &       0.08 \\
  IMDb &       2 &         40000 / 5000 &   6786 &           33.97 &       0.50 \\
  TAC-based RE &       41 &         1937211 / 18660 &   182292 &           0.51 &   -     \\
       \bottomrule
   \end{tabular}
   \caption[caption]{Summary of data statistics. The average rule hits are computed on the train set. Class ratio describes the amount of positive samples in the test set for binary classification datasets, i.e. data skewedness.}
   \label{tab:experiments_data_desc}
\end{table*}

Apart from denoising methods, \knodle{} includes a few datasets from previous works in the \knodle{}-specific tensor format in order to demonstrate the abilities of the framework.
All datasets are rather simple, but have their own peculiarities with respect to the respective $Z$ and $T$ matrices, that are worth investigating.
The overview of dataset statistics is provided in Table~\ref{tab:experiments_data_desc}.

\textbf{Spam Dataset.}
The first task uses the YouTube comments dataset~\cite{7424299}.
Here, the task is to classify whether a text is relevant to the video or holds spam, such as advertisement.
The dataset has a small size of both train and test sets.
Thus, a single wrongly labeled instance might have quite a big impact on the learning algorithm.
We use the preprocessed version by the Snorkel team~\cite{web:snorkel_spam}.
Among others, the rules were created based on keywords and regular expressions.

\textbf{Spouse Dataset.}
This relation extraction dataset is based on the Signal Media One-Million News Articles Dataset~\cite{Corney2016WhatDA}.
The task is to decide whether a sentence holds a spouse relation or not.
Again, the preprocessed version by the Snorkel team is used~\cite{web:snorkel_spouse}, so the results can be related to previous studies~\cite{snorkel}.
The rules are created via a set of known spouse relationships from DBPedia~\cite{DBPedia} as well as keywords and encoded language patterns.
The difficulty of the Spouse dataset is its skewness: over $90\%$ of samples in the test set hold a no-spouse relation.

\begin{table*}      
    \centering
    \begin{threeparttable}
        \captionsetup{font=10pt}
        \begin{tabular}{c c l c c c l c l c c c }
            \toprule
            &  \multicolumn{1}{c}{Spam} && \multicolumn{3}{c}{Spouse} && \multicolumn{1}{c}{IMDb} && \multicolumn{3}{c}{TAC-based RE} \\
            \cmidrule(lr){2-2} \cmidrule(lr){4-6} \cmidrule(lr){8-8} \cmidrule(lr){10-12}
            Mode & Acc && P & R & F1 && Acc && P & R & F1 \\
            \midrule [0.2ex]
            Majority vote & 0.81 && 0.12 & 0.79 & 0.22 && 0.65 && 0.09 & 0.001 & 0.001\\ [0.5ex]
            Majority + DistilBert & 0.87 && 0.09 & 0.90 & 0.17 && 0.67 && 0.20 & 0.19 & 0.19 \\ [0.5ex]
            \textit{k}-NN  + DistilBert & \textbf{0.94} && 0.12 & 0.86 & 0.21 && 0.50 && 0.10 & 0.11 & 0.10 \\ [0.5ex]
            Snorkel + DistilBert & 0.88 && 0.13 & 0.70 & \textbf{0.23} && 0.50 && - & - & - \\[0.5ex]
            WSCrossWeigh + DistilBert & \textbf{0.94} && 0.09 & 0.69 & 0.16 && \textbf{0.73} && 0.25 & 0.27 & \textbf{0.26} \\ [0.5ex]
            \bottomrule
        \end{tabular}
        \end{threeparttable}
        \caption{
                Results of the classifier training with different denoising methods on the test sets of datasets included in \textit{Knodle}. \newline
                \textsuperscript{\textdagger}The neighbors were searched with Approximate Nearest Neighbors~\cite{annoy} because of computation complexity of \textit{k}-NN search.
        }

        \label{tab:table}
\end{table*}

\textbf{IMDb Dataset.}
The third dataset is based on the well-known IMDb dataset~\cite{imdb_dataset}, which consists of short movie reviews.
The task is to determine whether a review holds a positive or negative sentiment.
Despite the training set has labels, we do not use them in our experiments, but handle this data in an unsupervised fashion.
To create the $Z$ and $T$ matrices, we use positive and negative keyword lists~\cite{10.1145/1014052.1014073}, with a total of $6800$ keywords.

\textbf{TAC-based Relation Extraction Dataset.}
Lastly, given the importance of distant supervision for relation extraction, we add a larger dataset with more relations (than just spouse).
For development and test purposes the TACRED corpus annotated via crowdsourcing and human labeling from KBP~\cite{zhang2017tacred} is used.
As human labels are not allowed in weak training, the training is performed not on the TACRED dataset, but on a weakly-supervised noisy corpus built on TAC KBP corpora~\cite{Surdeanu, Roth_2014}, which was annotated with entity pairs extracted from Freebase~\cite{freebase} with corresponding relations mapped to the 41 TAC relations.
The amount of entity pairs per relation is limited to 10.000 and each entity pair is allowed to be mentioned in no more than 500 sentences.
An important difference of this dataset to the other three is the presence of negative instances added to the dataset in equal proportion to the positive ones.

%% file: chapters/6_experimental_setup.tex
\section{Experiments}
\label{sec:experiments}

The aim of \knodle{} is not to find the best denoising method in general. Rather, the goal is to find the method that improves weak labels most for a given task or dataset and its specific properties.
Thus, \knodle{} supports experimentation to get a better understanding in which settings a certain method works well and when it does not.


\subsection{Experimental Details}

In all experiments, the DistilBert uncased model for English language~\cite{Sanh2019DistilBERTAD} provided by the HuggingFace \footnote{https://huggingface.co/}~\cite{wolf2020huggingfaces} library is used as the prediction model.
The optimization is performed with the AdamW optimizer~\cite{loshchilov2019decoupled} and a learning rate of $1e{-4}$.
We employ a cross-entropy loss accepting a probability distribution over all labels as reference input whenever the output of a denoising algorithm is a distribution over weak labels (e.g. \mbox{\texttt{kNNAggregationTrainer}}, \mbox{\texttt{SnorkelTrainer}}).
Reducing this representation to a single label (i.e.\ log-likelihood) would lead to a loss of weak signals, whereas a label distribution allows to exploit the information from $Z$ and $T$ to the fullest.
Each model was trained for 2 epochs (unless stated otherwise), which was enough to receive a stable result.

For the \textit{k}-NN algorithm, nearest neighbors were found using the cosine similarity of TF-IDF features based on a dictionary of $3000$ words, and the number of $k$ neighbors is treated as a hyper-parameter.
In our experiments, we used $k=2$ except where otherwise noted.
Hyperparameters for the WSCrossWeigh denoising algorithm are the number of folds the data is be split into, the number of partitions (that is, how many times the splitting for mistake estimation is done) and a weight-reducing rate (the value, by which the initial sample weights are reduced to each time the sample is predicted wrongly).
These parameters are tuned for each dataset individually.
The following best parameter values were found empirically: ($folds=3$, $partitions=10$ and $\epsilon=0.3$) for the Spam dataset, ($3$, $2$ and $0.3$) for the Spouse dataset and ($2$, $25$, $0.7$) for the IMDb dataset.
Apart from that, \knodle{} provides the opportunity to train the cross-validated sample weights with a model different from the final classifier.
In our experiments, the weights were calculated using a Bidirectional LSTM with GloVe Embeddings~\citep{pennington2014glove}, while the final training was performed with DistilBert using the same settings as in the experiments with other denoising methods.
The only difference is the number of epochs on the TAC-based dataset: the best results were obtained with $1$ DistilBert epoch.

\subsection{Results}
An overview of the results is given in Table~\ref{tab:table}.
In the Spam dataset, all denoising methods show an improvement over the simple majority vote baseline.
The data-adaptive \textit{k}-NN and WSCrossWeigh methods perform best in this setting.
Snorkel and standard majority voting followed by DistilBert fine-tuning overfit to the noisy majority votes.
This becomes obvious with the observation that Snorkel achieves a score of $0.93$ with a simple logistic regression discriminative model.
Interestingly, \textit{k}-NN performs well which can serve as a proof for the reliability of neighboring labels.

Compared to the Spam dataset, the Spouse dataset is much larger.
As the task is to find sentences holding spouse relations, we relate all metrics to the \textit{is-spouse} relation.
Note that the \textit{non-spouse} relation remains in this case completely disregarded. 
Furthermore, the class ratio equals $0.08$ shows that \textit{is-spouse} is the complicated class of interest. 
On average, $0.34$ rules hit per instance, meaning that almost $70\%$ of the data match no rule.  In these cases, majority vote uses a random vote which oversamples the \textit{is-spouse} relation, rendering a high recall but low precision.
We found that the rule matches overrepresent the \textit{is-spouse} class as they are closer to a class ratio of $0.5$ than to the true class ratio of $0.9$.
Thus, the additional model training magnifies overfitting towards the \textit{is-spouse} class which, again, is expressed by increased recall and lower precision.
The only denoising system that generalizes is Snorkel.
One possible explanation could be that it is the only method that provides explicit rule denoising.

For IMDb, the majority vote shows that the rules have rather low quality on their own, but an additional trained model on top manages to generalize beyond the given labels.
In contrast, denoising with the \textit{k}-NN algorithm only aggravates the problems inherent to labels as the classifier's performance drops down to a random vote ($50\%$ accuracy).
This behaviour can be explained by the high density of rule hits: on average, no less than $33$ keywords match for each sentence,
which means that already for $k=1$ many neighbors are added and that the propagation of imprecise labelings overrules the expected benefits of $k$-NN\@.
In general, there are cues that \textit{k}-NN might useful in cases where the weak labels are already rather reliable but fail in cases where weak labels are too noisy.
The Snorkel based denoising does not perform well on IMDb dataset as well, which can be explained by the lack of dependencies between the rules that the Snorkel system relies on.
However, WSCrossWeigh appears to be very robust to these data characteristics, the large amount of rules seems to help tease out and mutually reinforce the data characteristics associated with a specific label in cross-validation.

The distantly supervised TAC-based RE dataset turns out to be the most complicated dataset among all because
of a larger size of samples $n$ and a larger number of rules $r$.
Due to its specificity, there are almost no rule matches (entity pairs from the seed KB) on the test set, implying that the simple majority baseline has scores close to $0$.
Training with DistilBert improves the result, however the performance remains considerably worse than for the data sets discussed above.
On the contrary the WSCrossWeigh method that not directly denoise the rules, but downweigh the mislabeled data samples is still able to improve the results.
Snorkel denoising could not be performed on this dataset on a machine with CPU frequency of 2.2GHz with 40 cores due to the immense amount of rules without the data manipulations we want to avoid (such as significantly reducing the number of rules).
The computation of distances between almost 2 millions instances, which are necessary to determine the nearest neighbors, also turned out to be extremely memory- and time-consuming, explaining why \textit{k}-NN algorithm was also not performed.
Instead, we work around this by applying an \textit{approximated} \textit{k}-NN algorithm.
In our experiments we used the Annoy library~\cite{annoy} and $k=3$ parameter.
The poor performance of approximated $k$-NN could be explained by a small average of rule hits in the TAC-based RE data set;
the possible approximation losses are also not to be neglected.
In contrast, the WSCrossWeigh method performs quite well.
Our explanation is that WSCrossWeigh does not directly denoise the rules, but down-weighs samples it is less confident about.
This makes this approach more robust in cases where the rules are very noisy.



%% file: chapters/7_conclusion.tex
\section{Conclusion}

This work introduces the Knowledge-supervised Deep Learning framework \knodle{}.
\knodle{} provides a unified interface to work with multiple weak labeling sources, so that they can be seamlessly integrated with the training of deep neural networks.
This is achieved by a tensor-based input format and a intuitive separation of weak supervision aspects and model training.
The framework facilitates experimentation that helps researchers to gain better insights into the correspondence between characteristics of weak supervision problems, and the effectiveness of methods for improving weakly supervised learning.
From a practical perspective, \knodle{} can be used to compare different denoising methods and select the one that gives the best result for a specific task.

\knodle{}'s modular approach makes it easy to add new data sets and denoising algorithms.
Adding functionality to \knodle{} is straightforward, and we do hope  
that it will encourage researchers to create their own algorithms to improve learning with weakly annotated data, and incorporate them into the \knodle{} framework.

%% file: chapters/8_acknowledgements.tex
\section*{Acknowledgments}

This research was funded by the WWTF through the project "Knowledge-infused Deep Learning for Natural Language Processing" (WWTF Vienna Research Group VRG19-008), by the Deutsche Forschungsgemeinschaft (DFG, German Research Foundation) - RO 5127/2-1, and supported by a gift from Diffbot \footnote{\url{https://www.diffbot.com/}}.